\documentclass{sig-alternate}
\pdfoutput=1 
\usepackage{mathptmx} 

\usepackage{fancyhdr}
\usepackage[normalem]{ulem}
\usepackage[hyphens]{url}
\usepackage[square,sort,comma,numbers]{natbib}
\usepackage[protrusion=true,tracking=true,kerning=true,expansion,final]{microtype}  
\usepackage{flushend}

\usepackage{amsfonts}
\usepackage{physics} 
\usepackage{wrapfig}    
\usepackage{xcolor} 
\usepackage{booktabs} 
\usepackage{nicefrac} 
\usepackage{cancel}
\usepackage{gensymb}
\usepackage{svg}
\usepackage{bbding}
\usepackage{mathtools}
\usepackage{centernot} 
\usepackage{bigints}
\usepackage{dsfont} 
\usepackage{esint} 
\usepackage{multirow}
\usepackage{lipsum} 
\usepackage{ulem}
\usepackage{verbatim}
\usepackage{wrapfig}
\usepackage{svg}
\usepackage{amsmath}

\usepackage{varioref} 

\usepackage[bookmarks=true,breaklinks=true,letterpaper=true,colorlinks,linkcolor=black,citecolor=blue,urlcolor=black]{hyperref}

\hypersetup{ %
	pdftitle={HCM: Hardware-Aware Complexity Metric for Neural Network Architectures}, 
	pdfauthor={Alex Karbachevsky, Chaim Baskin, Evgenii Zheltonozshkii, Yevgeny Yermolin, Freddy Gabbay, Alex M. Bronstein, and Avi Mendelson},
	pdfsubject={},
	pdfkeywords={},
	pdfborder=0 0 0,
	pdfpagemode=UseNone,
	pdfview=FitH}
	
\usepackage[capitalise]{cleveref} 
\crefname{appsec}{appendix}{appendices}
\Crefname{appsec}{Appendix}{Appendices}

\vbadness=\maxdimen
\title{HCM: Hardware-Aware Complexity Metric \\ for Neural Network Architectures}

\newcommand*\samethanks[1][\value{footnote}]{\footnotemark[#1]}
\author{
Alex Karbachevsky\,${^\dagger}$\thanks{Equal contribution.}\quad 
Chaim Baskin\,${^\dagger}$\samethanks[1]\quad 
Evgenii Zheltonozshkii\,$^\dagger$\samethanks[1]\quad
Yevgeny Yermolin\,$^\dagger$
\\[0.15cm]
Freddy Gabbay\,$^\circ$\quad 
Alex M. Bronstein\,$^\dagger$\quad 
Avi Mendelson\,$^\dagger$
\\[0.2cm]
$^\dagger$Technion -- Israel Institute of Technology, Haifa, Israel \\
$^\circ$Ruppin Academic Center, Haifa, Israel,\\[0.2cm]
\small{\texttt{\{\href{mailto:alex.k@campus.technion.ac.il}{alex.k},  \href{mailto:chaimbaskin@campus.technion.ac.il}{chaimbaskin},  \href{mailto:evgeniizh@campus.technion.ac.il}{evgeniizh}\}@campus.technion.ac.il}}\\
\small{\texttt{\{\href{mailto:yevgeny.ye@cs.technion.ac.il}{yevgeny.ye},  \href{mailto:bron@cs.technion.ac.il}{bron}, \href{mailto:avi.mendelson@cs.technion.ac.il}{avi.mendelson}\}@cs.technion.ac.il}}\\
\small{\texttt{\{\href{mailto:freddyg@ruppin.ac.il}{freddyg}\}@ruppin.ac.il}}
}

\usepackage{float}

\begin{document}
\maketitle
\pagestyle{plain}


\begin{abstract}
Convolutional  Neural  Networks  (CNNs)  have become common in many fields  including computer vision, speech recognition, and natural language processing. Although CNN hardware accelerators are already included as part of many SoC architectures, the task of achieving high accuracy on resource-restricted devices is still considered challenging, mainly due to the vast number of design parameters that need to be balanced to achieve an efficient solution. Quantization techniques, when applied to the network parameters, lead to a reduction of power and area and may also change the ratio between communication and computation. As a result, some algorithmic solutions may suffer from lack of memory bandwidth or computational resources and fail to achieve the expected performance due to hardware constraints. Thus, the system designer and the micro-architect need to understand at early development stages  the impact of their high-level decisions (e.g., the architecture of the CNN and the amount of bits used to represent its parameters) on the final product (e.g., the expected power saving, area, and accuracy). Unfortunately, existing tools fall short of supporting such decisions. 

This paper introduces a hardware-aware complexity metric
that aims to assist the system designer of the neural network architectures, through the entire project lifetime  (especially at its early stages) by predicting the impact of architectural and micro-architectural decisions on the final product. 
We demonstrate how the proposed metric can help evaluate different design alternatives of neural network models on resource-restricted devices such as real-time embedded systems, and to avoid making design mistakes at early stages. 

\end{abstract}

\section{Introduction}
\label{sec:intro}

\begin{figure}
 \centering
        \includegraphics[width=0.8\linewidth]{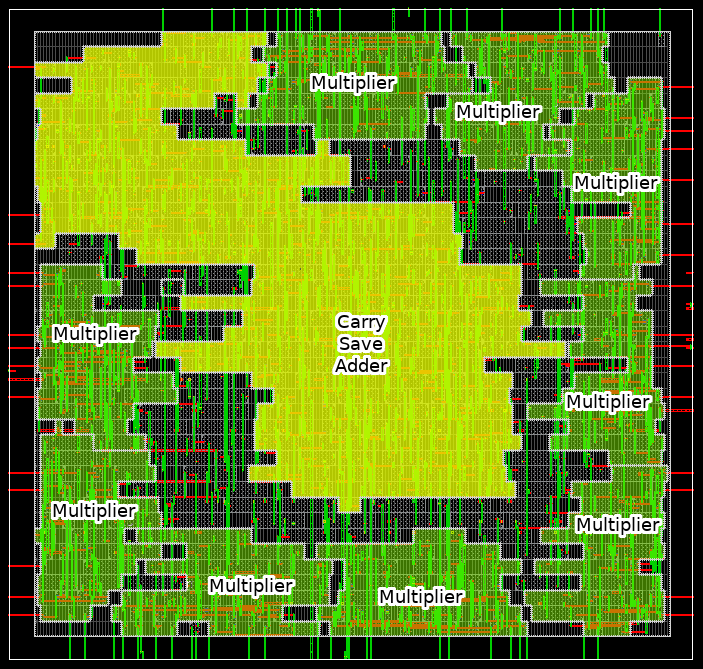}
        \caption{Our $3\times3$ kernel 8-bit processing engine (PE) layout using the TSMC 28nm technology. The carry-save adder can fit 12-bit numbers, which is large enough to store the output of the convolution. }
        \label{fig:PE_layout}
\end{figure}

Domain-specific systems were found to be very efficient, in general, and when developing constrained devices such as IoT, in particular. A system architect of such devices must consider hardware limitations (e.g., bandwidth and local memory capacity), algorithmic factors (e.g., accuracy and representation of data), and system aspects (e.g., cost, power envelop, battery life, and more).  Many IoT and other resource-constrained devices  provide support for applications that use convolutional neural networks (CNNs). Such algorithms can achieve spectacular performance in various tasks covering a wide range of domains such as computer vision, medicine, autonomous vehicles, etc. Notwithstanding, CNNs contain a vast number of parameters and require a significant amount of computation during inference, thus monopolizing hardware resources and demanding massively parallel computation engines; see teh example shown in \cref{fig:PE_layout}. 

These requirements have led to great interest in using custom-designed hardware for efficient inference of CNNs that would allow the promise of neural networks to be used in real-life applications by deploying them on low-power edge devices.
Developing such systems requires a new set of design tools due to the tight entanglement between the algorithmic aspects, the chip architecture and the constraints the end product needs to meet. 
In particular, great efforts were made to develop low-resource CNN architectures  \cite{Howard_2019,Wu2018FBNetHE,Sandler2018MobileNetV2IR,ridnik2020tresnet}.
One example of such architectural changes is the splitting of the regular $3\times3$ convolutions into a channel-wise $3\times3$ convolution followed by a $1\times1$ one. Another way to reduce the computational burden  is to quantize the CNN parameters, weights and activations, employing low-bit integer representation of the data instead of the expensive floating point representation. Recent quantization-aware training schemes \cite{esser2019learned,zhao2020linear,gysel2016hardware,yang2019quantization,jin2019towards} achieve near-baseline accuracy for as low as 2-bit quantization. The benefit of quantizing the CNN is twofold: both the number of gates required for each multiply-accumulate (MAC) operation and the amount of routing are reduced.  The decision regarding which algorithm to choose may depend on the architecture (e.g., FPGA or ASIC), the accuracy requirements, and their impact on performance and power. Thus, the architect needs to make these fundamental decisions early in the developing process and no existing tool can help predict these design factors ahead of time. 

The impact of the high-level structure of the accelerator, e.g., the type of CNN levels and the representation of the  operands, on the power, the area and the performance of the final product needs to be defined and predicted at an early stage of the project.
Recent research has shown that ASIC-based architectures are the most efficient solution for CNN accelerators both in datacenters \cite{jouppi2017datacenter,raihan2019modeling,chen2019eyerissv2} and in real-time platforms \cite{eyeris,han2016eie,Rivas_Gomez_2018}. Accordingly, we demonstrate the proposed metric and design tool on an implementation of a streaming \cite{reddi2019mlperf} ASIC-based convolutional engine. Nevertheless, our methodology can be applied for the evaluation of other types of architectures, such as FPGA-based accelerators \cite{ankit2019puma,baskin2018streaming,10.1145/3020078.3021744}. In both cases, the development process includes an important trade-off between the logical gates area, their routing on the silicon and the performance of the resulting system. Unfortunately, all these parameters also depend on the representation of the data, and its impact on both communication and computation. To date, there is no quantitative metric for this trade-off available at the design stage of the CNN accelerator and no tool exists that can assist the architect to predict the impact of high level decisions on the important design parameters. Ideally, the designer would like to have an early estimation of the chip resources required by the accelerator as well as the performance, accuracy and power it can achieve.

A critical difficulty in trying to predict the design parameters for CNN-based systems is the lack of a proper complexity metric. Currently, the most common metric for calculating the computational complexity of CNN algorithms is the number of MAC operations denoted as OPS (or FLOPS in case of floating-point operations). This metric, however, does not take into account the data format or additional operations performed during the inference, such as memory accesses and communication. For that reason, the number of FLOPS does not necessarily correlate with runtime \cite{lee2020compounding} or the required amount of computational resources.
This paper proposes to use a different metric for assessing the complexity of CNN-based architectures: the number of bit operations (BOPS) as presented by \citet{baskin2018uniq}. We show that BOPS is well-suited to the task of comparing different architectures with different weight and activation bitwidths. 

\paragraph{Contribution} This paper makes the following contributions: Firstly, we study the impact of CNN quantization on the hardware implementation in terms of computational resources and memory bandwidth considerations. Specifically, we study a single layer in the neural network.

Secondly, we extend the previously proposed computation complexity for quantized CNNs, termed BOPS \cite{baskin2018uniq}, with a communication complexity analysis to identify the performance bottlenecks that may arise from the data movement. 

Thirdly, we extend the roofline model \cite{williams2009roofline} to accommodate this new notion of complexity. We also demonstrate how this tool can be used to assist architecture-level decisions at the early design stages.

Lastly, we implement a basic quantized convolution block with various bitwidths on 28nm processes to demonstrate an accurate estimation of the power/area of the hardware accelerator. This allows changing high-level design decisions at early stages and saves the designer from  major mistakes that otherwise would be discovered too late. We compare our estimations with previous approaches and show  significant improvement in accuracy of translation between algorithmic complexity to hardware resource utilization.

The rest of the paper is organized as follows: \cref{sec:related} reviews the related work; \cref{sec:method} describes a proposed hardware-aware complexity metric; \cref{sec:roofline} provides roof-line analysis of CNN layer design using the proposed metric; \cref{sec:expirements} provides the experimental results using common CNN architecture and \cref{sec:conclussions} concludes the paper.

\section{Related work}
\label{sec:related}
In this section, we provide an overview of prior work that proposed metrics for estimating the complexity and power/energy consumption of different workloads, focusing on neural networks. The most commonly used metric for evaluating computational complexity is FLOPS \cite{mcmahon1986livermore}: the amount of floating point operations required to perform the computation. In the case of integer operations, the obvious generalization of FLOPS is OPS, which is just the number of operations. A fundamental limitation of these metrics is the assumption that the same data representation is used for all operations; otherwise, the calculated complexity does not reflect the real one.  
\citet{Wang2018BOPSNF} claim that FLOPS is an inappropriate metric for estimating the performance of workloads executed in datacenters and proposed a basic operations metric that uses a roofline-based model, taking into account the computational and communication bottlenecks for more accurate estimation of the total performance.

In addition to general-purpose metrics, other metrics were developed specifically for evaluation of neural network complexity. 
\citet{mishra2018wrpn} define  the ``compute cost'' as the product of the number of fused multiply–add (FMA) operations and the sum of the width of the activation and weight operands, without distinguishing between floating- and fixed-point operations. Using this metric, the authors claimed to have reached a $32\times$ ``compute cost'' reduction  by switching from FP32 to binary representation. Still, as we show further in our paper, this is a rather poor estimate for the hardware resources/area needed to implement the computational element. 
\citet{jiang2019pitfall} notes that a single metric cannot comprehensively reflect the performance of deep learning (DL) accelerators. They investigate the impact of various frequently-used hardware optimizations on a typical DL accelerator and quantify their effects on accuracy and throughout under-representative DL inference workloads. Their major conclusion is that high hardware throughput is not necessarily highly correlated with the end-to-end high inference throughput
of data feeding between host CPUs and AI accelerators.
Finally, \citet{baskin2018uniq} propose to generalize FLOPS and OPS by taking into account the bitwidth of each operand as well as the operation type. The resulting metric, named BOPS (binary operations),  allows area estimation of quantized neural networks including cases of mixed quantization. 

The aforementioned metrics do not provide any insight  on the amount of silicon resources needed to implement them. Our work, accordingly, functions as a bridge between the CNN workload complexity and the real power/area estimation. 

\section{Complexity Metric}
\label{sec:method}
In this section, we describe our hardware-aware complexity metric (HCM), which takes into account the CNN topology,  and define the design rules of efficient implementation of quantized neural networks. The HCM metric assesses two elements: the computation complexity, which quantifies the hardware resources needed to implement the CNN on silicon, and the communication complexity, which defines the memory access pattern and bandwidth. We describe the changes resulting from switching from a floating-point representation to a fixed-point one, and then present our computation and communication complexity metrics.
All results for the fixed-point multiplication presented in this section are based on the Synopsys standard library multiplier using TSMC's 28nm process.

\subsection{The impact of quantization on hardware implementation}
\label{sec:quantization}

\begin{table*}
	\centering
	\caption{32-bit floating-point and 32-bit fixed-point multiplier hardware complexity in terms of the number of gates, area, and power.}
	\label{table:float_fixed_mult_area}
	\begin{center}
		\begin{small}
			\begin{tabular}{lccccccc}
				\toprule
				\multirow{2}{*}{\bf{Multiplier}} & \multirow{2}{*}{\bf{Gates}} & \multirow{2}{*}{\bf{Cells}} & \multirow{2}{*}{\bf{Area $[\mu m^2]$}} & \multicolumn{4}{c}{\bf{Power$[mW]$} }     \\
				&  &  &  & \bf{Internal } & \bf{Switching } & \bf{Leakage } & \bf{Dynamic }       \\
				\midrule
				Floating-Point & 40090 & 17175 & 11786 & 2.76 & 1.31 & 0.43 & 10.53 \\
				Fixed-Point & 5065 & 1726 & 1489 & 0.49 & 0.32 & 0.04 & 1.053 \\
				\bottomrule
			\end{tabular}
		\end{small}
	\end{center}
\end{table*}

Currently, the most common representation of weights and activations for training and inference of CNNs is either 32-bit or 16-bit floating-point numbers. The fixed-point MAC operation, however, requires significantly fewer hardware resources, even for the same input bitwidth. To illustrate this fact, we generated two multipliers: one for 32-bit floating-point\footnote{FPU100 from \url{https://opencores.org/projects/fpu100}} and the other for 32-bit fixed-point operands. The results in \cref{table:float_fixed_mult_area} show that a fixed-point multiplier uses approximately eight time less area, gates, and power than the floating-point counterpart. Next, we generated a convolution with a $k \times k$ kernel, a basic operation in CNNs consisting of $k^2$ MAC operations per output value. After switching from floating-point to fixed-point, we explored the area of a single processing engine (PE) with variable bitwidth. Note that accumulator size depends on the network architecture: the maximal bitwidth of the output value is $b_wb_a + \log_2(k^2)+\log_2(n)$, where $n$ is number of input features. Since the extreme values are very rare, however, it is often possible to reduce the accumulator width without harming the accuracy of the network \cite{chen2019eyerissv2}.

\begin{figure}
 \centering
        \includegraphics[width=\linewidth]{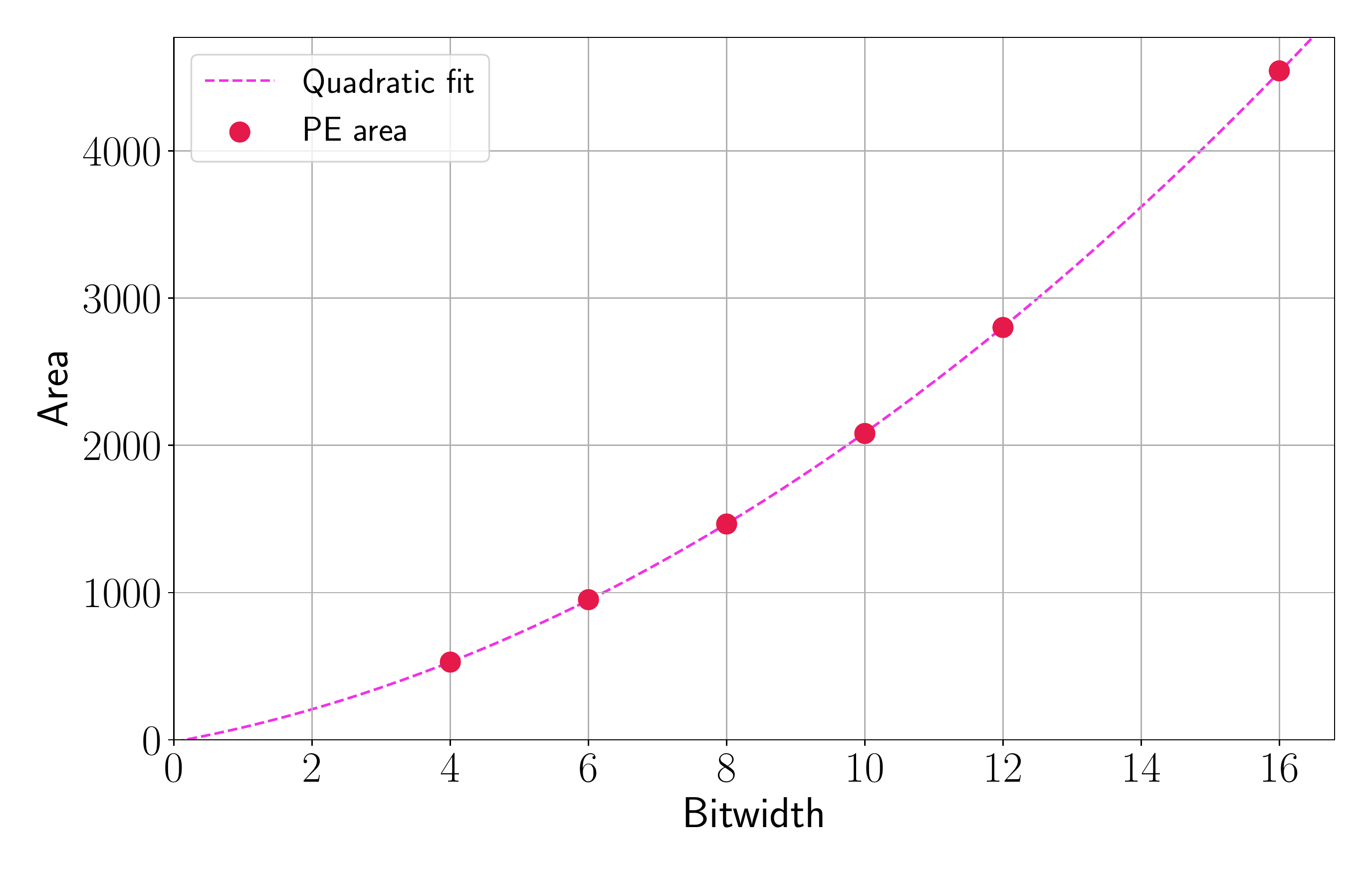}
        \caption{Area ($A$) vs. bitwidth ($b$)  for a $3\times 3$ PE with a single input and output channel. All weights and activations use the same bitwidth and the accumulator width is 4 bit larger, which is enough to store the result. The quadratic fit is $A = 12.39b^2+ 86.07b-14.02$ with goodness of fit $R^2=0.9999877$.     }
        \label{fig:bits_vs_area}
\end{figure}

\cref{fig:bits_vs_area} shows the silicon area of the PE as a function of the bitwidth. We performed a polynomial regression and observed a quadratic dependence of the PE area on the bitwidth, with the coefficient of determination $R^2=0.9999877$. This nonlinear dependency demonstrates that quantization impact a network hardware resources is quadratic: reducing bitwidth of the operands by half reduces area and, by proxy, power approximately by a factor of four (contrary to what is assumed by, e.g., \citet{mishra2018wrpn}).

\subsection{Computation}
\label{sec:Comp}
We now present the BOPS metric defined in \citet{baskin2018uniq} as our computation complexity metric. In particular, we show that BOPS can be used as an estimator for the area of the accelerator. The area, in turn, is found to be linearly related to the power in case of the PEs. 

The computation complexity metric describes the amount of arithmetic ``work'' needed to calculate the entire network or a single layer. BOPS is defined as the number of bit operations required to perform the calculation: the multiplication of $n$-bit number by $m$-bit number requires $n\cdot m$ bit operations, while addition requires $\max(n,m)$ bit operations.  In particular,  \citet{baskin2018uniq} show that a $k\times k$ convolutional layer with $b_a$-bit activations and $b_w$-bit weights requires 
\begin{align}\label{BOPS_eq}
    \mathrm{BOPS} &= m n k^2 \qty(b_a b_w + b_a+b_w+\log_2(n k^2))
\end{align}
bit operations, where $n$ and $m$ are, respectively, the number of input and output features of the layer. The formula takes into account the width of the accumulator required to accommodate the intermediate calculations, which depends on $n$. The BOPS of an entire network is calculated as a sum of the BOPS of the individual layers. 
Creating larger accelerators that can process more layers in parallel involves simply replicating the same individual PE design.

\begin{figure}
 \centering
        \includegraphics[width=1.0\linewidth]{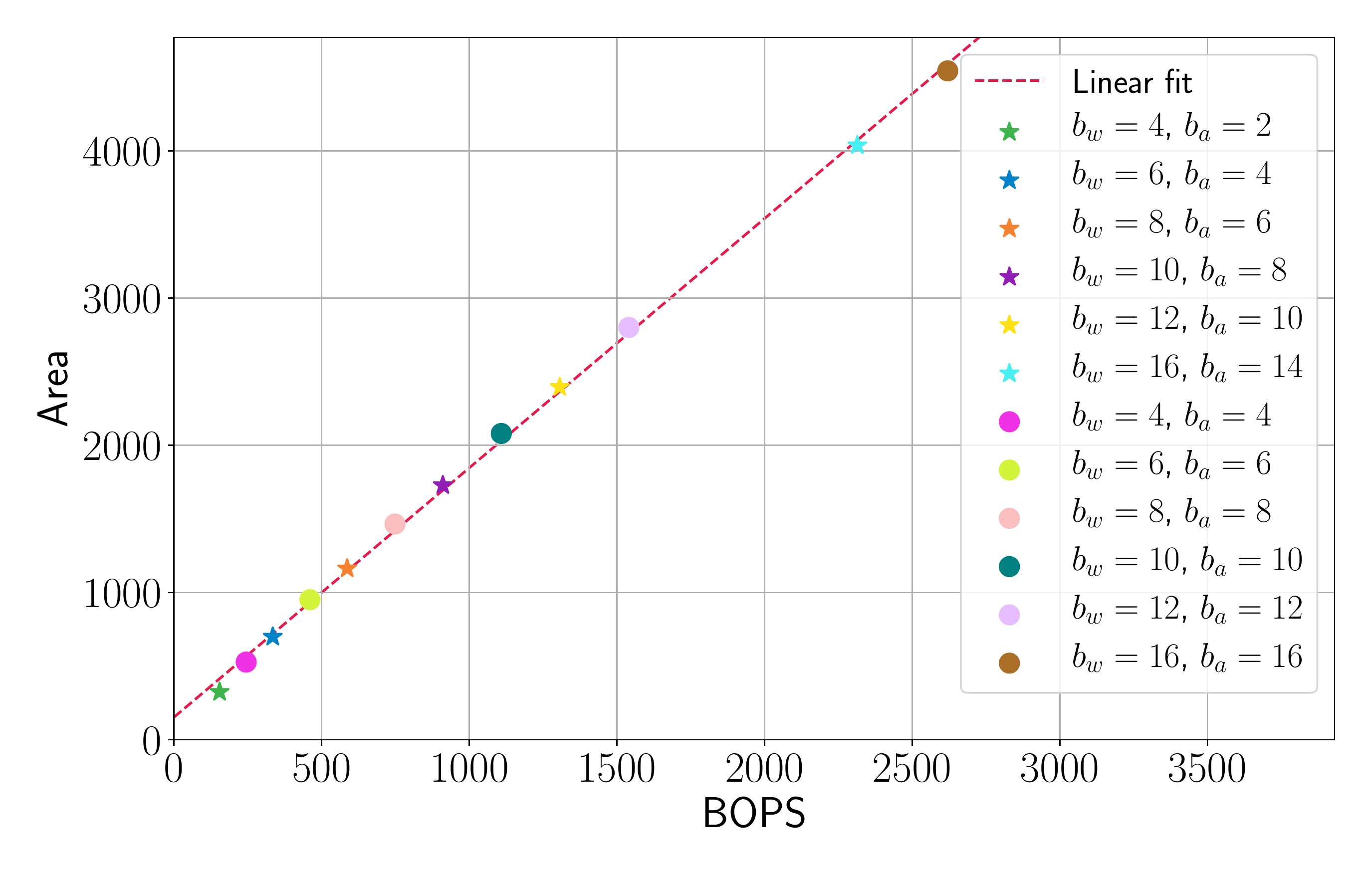}
        \caption{Area ($A$) vs. BOPS ($B$) for a $3 \times 3$ PE with a single input and output channel and variable bitwidth. The linear fit is $A=1.694B+153.46$ with goodness of fit $R^2=0.9989$.}
        \label{fig:area_vs_BOPS_1x1}
\end{figure}

\begin{figure}
 \centering
        \includegraphics[width=1.0\linewidth]{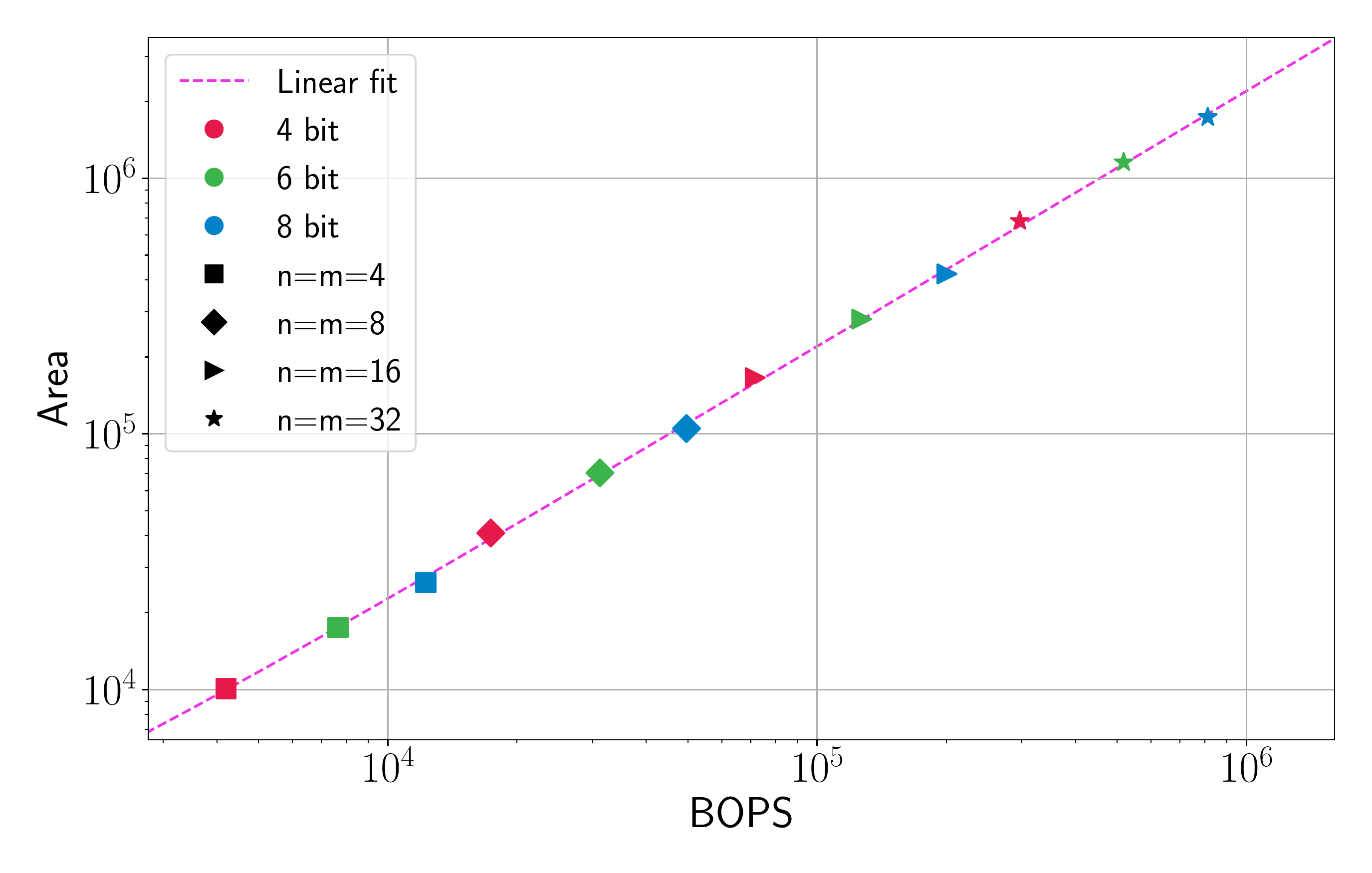}
        \caption{Area ($A$) vs. BOPS ($B$) for a $3 \times 3$ PE with variable input ($n$) and output ($m$) feature dimensions, and variable bitwidth. Weights and activations use the same bitwidth and the accumulator width is set to $\log_2(9m) \cdot b_w \cdot b_a$.}
        \label{fig:area_vs_BOPS4}
\end{figure}

In \cref{fig:area_vs_BOPS_1x1}, we calculated BOPS values for the PEs from \cref{fig:bits_vs_area} and plotted them against the area.
We conclude that for a single PE with variable bitwidth,  BOPS can be used to predict the PE area with high accuracy. 

Next, we tested the predictive power of BOPS scaling with the size of the design. We generated several designs with variable bitwidths, $b_w=b_a\in \qty{4,6,8}$, and variable numbers of PEs $n=m \in \qty{4,8,16}$ used to accommodate multidimensional inputs and outputs that typically arise in real CNN layers.  
\cref{fig:area_vs_BOPS4} shows that the area depends linearly on the BOPS for the range of two orders of magnitude of total area with goodness of fit $R^2=0.9980$. We conclude that since BOPS provides a high-accuracy approximation of the area and power required by the hardware, it can be used as an early estimator. While the area of the accelerator depends on the particular design of the PE, this only affects the slope of the linear fit, since the area is still linearly dependent on the amount of PEs. An architect dealing with algorithms only can use definitions such as the number of input features and output features, kernel size etc. and get an early estimation how much power is needed to solve the network, without having any knowledge about VLSI constraints in advance. Using information such as a number of input/output features and kernel size, it is possible  to immediately assess the amount of area the PEs occupy on the silicon.

\subsection{Communication}
\label{sec:Comm}
Another important aspect of hardware implementation of CNN accelerators is memory communication.
The transmission of data from the memory and back  is often overlooked by hardware implementation papers \cite{7551379,eyeris,ankit2019puma} that focus on the raw calculation ability to determine the performance of their hardware. In many cases, there is a difference between the calculated performance and real-life performance, since real-life implementations of accelerators are often memory-bound \cite{Morcel:2019:FAC:3322884.3306202,jouppi2017datacenter,wang2019lutnet}.

For each layer, the total memory bandwidth is the sum of the activation and weight sizes read and written from memory. In typical CNNs used, e.g., in vision tasks, the first layers consume most of their bandwidth for activations, whereas in deeper layers that have smaller but higher-dimensional feature maps (and, consequently, a bigger number of kernels), weights are the main source of memory communication.

\begin{figure}
 \centering
        \includegraphics[width=0.9\linewidth]{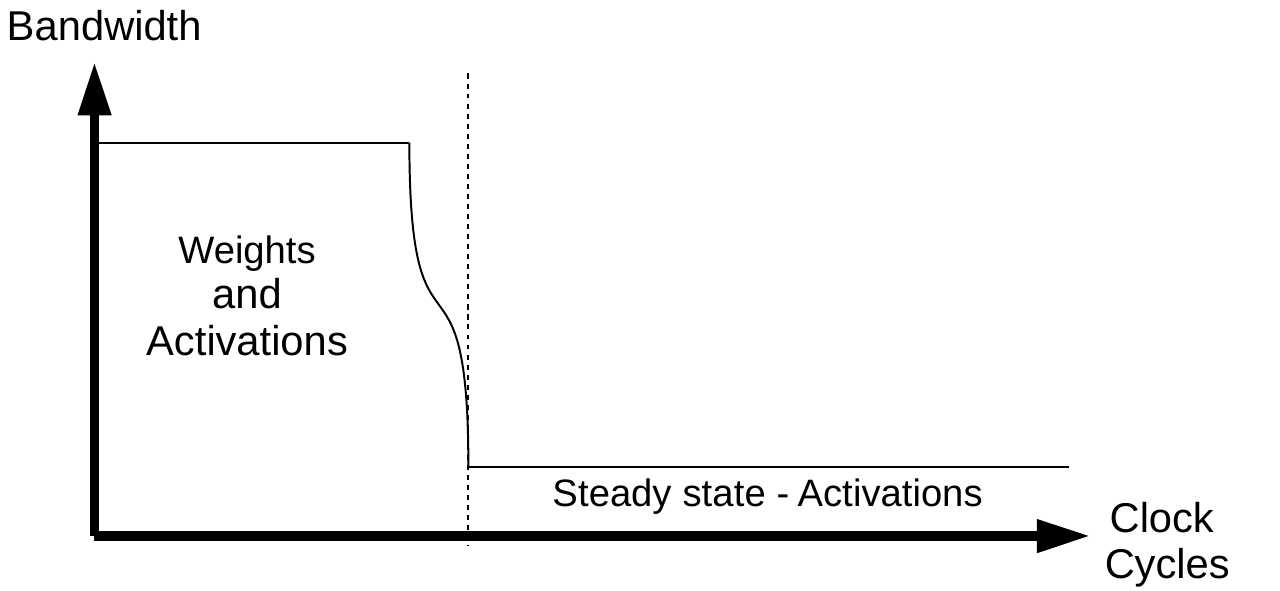}
        \caption{Per-layer memory access pattern}
        \label{fig:bw_per_clock}
\end{figure}

We assume that each PE can calculate one convolution result per clock cycle  and the resulting partial sum is saved in the cache. In \cref{fig:bw_per_clock}, we show typical memory access progress at the beginning of the convolutional layer calculation. At first stage, the weights and the first $k$ rows of the activations are read from memory at maximal possible speed to start the calculations as soon as possible. After the initial data are loaded, the unit reaches a ``steady state'', in which it needs to read from the memory only one new input value per clock cycle (other values are already in the cache). We assume the processed signals to be two-dimensional (images), which additionally requires $k$ new values to be loaded in the beginning of each new row. 

Note that until the weights and the first activations are loaded, no calculations are performed. The overhead bandwidth of the pre-fetch stage can be mitigated by doing work in greater batch sizes, loading the weights once and reading several inputs for the same weights. By doing this, we minimize the penalty for reading the weights compared to reading the actual input data to perform the calculation. In the case of real-time processing, however, larger batches are not possible because the stream of data needs to be completed on-the-fly. We focus on the latter real-time streaming regime in this paper because of its great importance in a range of applications including automotive, security, and finance.
The memory access pattern depicted in \cref{fig:bw_per_clock} must be kept in mind when designing the hardware, since it may limit the performance of the accelerator and decrease its power efficiency.

\section{Roofline analysis}
\label{sec:roofline}

\begin{figure}
    \centering  
    \includegraphics[width=1.0\linewidth]{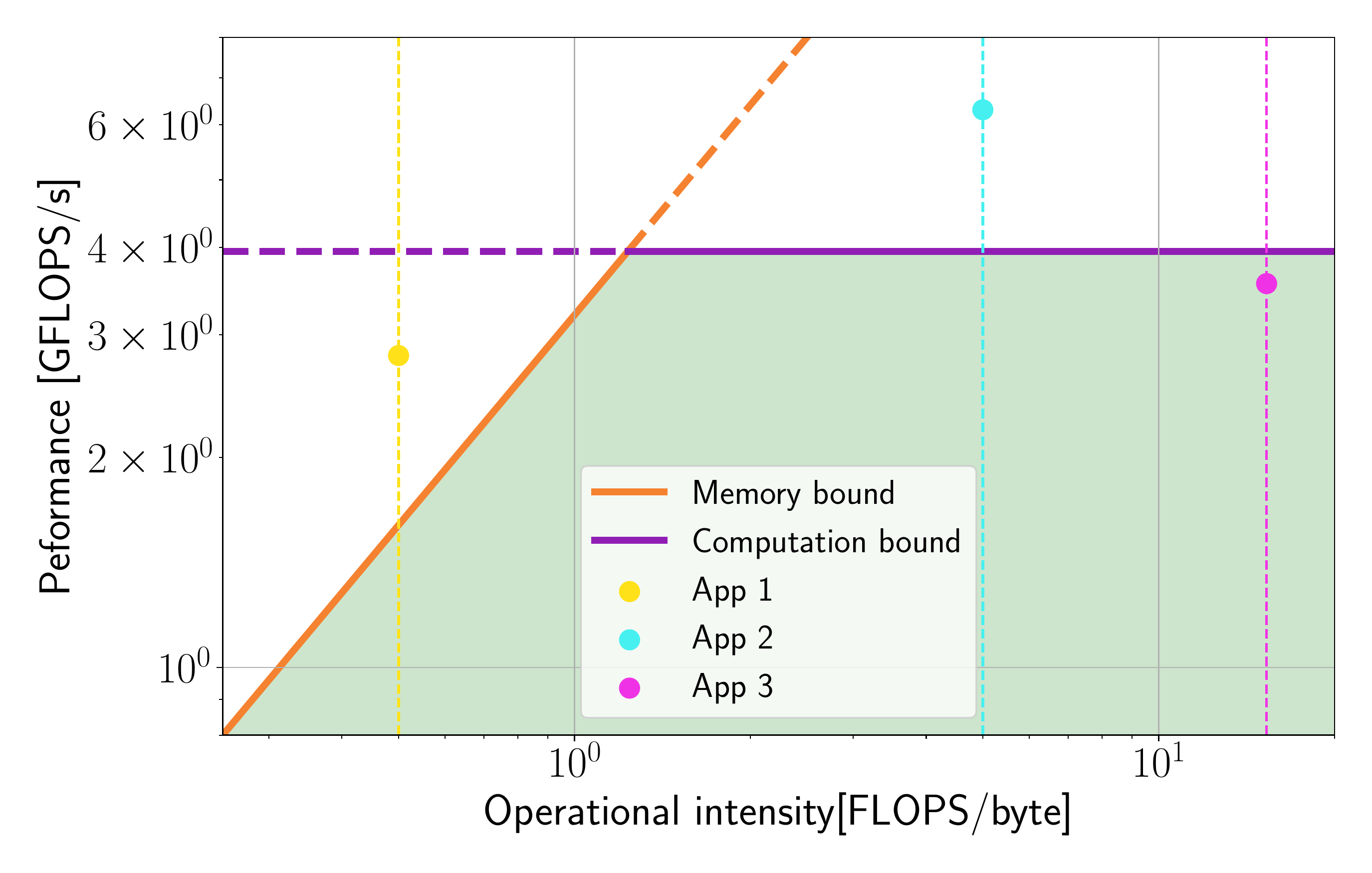}
    \caption{Roofline example. In the case of $\mathrm{App 1}$, memory bandwidth prevents the program from achieving its expected performance. In the case of $\mathrm{App 2}$, the same happens due to limited computational resources. Finally, $\mathrm{App 3}$ represents a program that could achieve its maximum performance on a given system.}
    \label{fig:basic_roofline}
\end{figure}

So far, we discussed the use of BOPS for the prediction of the physical parameters of the final product, such as the expected power and area. In this section, we extend the BOPS model  to a system level, by introducing the  OPS-based roofline model. 
The traditional roofline model, as introduced by \citet{williams2009roofline}, suggests depicting the dependencies between the performance (e.g., GFLOPS/second) and the operation density (the average number of operations per information unit transferred over the memory bus).  Now, for each machine we can draw ``roofs'': the horizontal line that represents its computational bounds and the diagonal line that represents its maximal memory bandwidth. An example of the roofline for three applications assuming infinite compute resources and memory bandwidth is shown in \cref{fig:basic_roofline}.   The maximum performance a machine can achieve for any application is visualized by the area below both bounds, shaded in green.  

\paragraph{OPS-based roofline model}
Since, as indicated in \cref{sec:quantization}, FLOPS cannot be used for efficient estimation of the complexity of quantized CNNs, we introduce a new model that is based on the BOPS metric presented in 
\cref{sec:Comp}. This model, to which we refer as the OPS-based roofline model, replaces the GFLOPS/s axis of the roofline plot with a performance metric more adequate for neural networks, e.g., number of operations per second (OPS/s), and the second metric that measures the computational complexity with operations per bit (OPS/bit). Using generic operations and bits allows plotting quantized accelerators with different bitwidths on the same plot.

As an example of the proposed approach,  we use two different ResNet-18 layers (a deep layer, which is computationally-intensive, and an early one, which is memory-intensive) on four different accelerator designs: 32-bit floating-point, 32-bit fixed-point, and quantized 8-bit and 4-bit fixed-point. The accelerators were implemented using standard ASIC design tools, as detailed in \cref{sec:expirements} and were built using the TSMC 28nm technology, using standard $2.4$GHz DDR-4 memory with a 64-bit data bus.

The first example employs an accelerator with a silicon area of $1\mathrm{mm^2}$ and 800MHz clock speed. The task is the 11$^{th}$ layer of ResNet-18 that has a $3\times3$ kernel and 256 input and output features of dimension $14\times 14$ each. Looking at \cref{table:float_fixed_mult_area}, it is possibly to fit only $\mathbf{85}$ 32-bit floating-point multipliers in $1\mathrm{mm^2}$. That allows installation of $\mathbf{9}$ PEs (without taking into account the area required for the accumulators of the partial sums) and calculation of convolutions with the $3\times 3\times 3\times 3$ kernel in a single clock. Using the known areas of 4-bit, 8-bit and 16-bit PEs, we extrapolate the area of the 32-bit fixed point PE to be $16676\mu \mathrm{m}$. From these data, we can place $\mathbf{60}$ PEs with $7\times 7\times 3\times 3$ kernels, $\mathbf{220}$ PEs with $14\times 14\times 3\times 3$ kernels and $\mathbf{683}$ PEs with $26\times 26\times 3\times 3$ kernels, for 32-bit, 16-bit and 8-bit fixed-point PEs, respectively, on the given area. 

\begin{figure}
 \centering
        \includegraphics[width=1.0\linewidth]{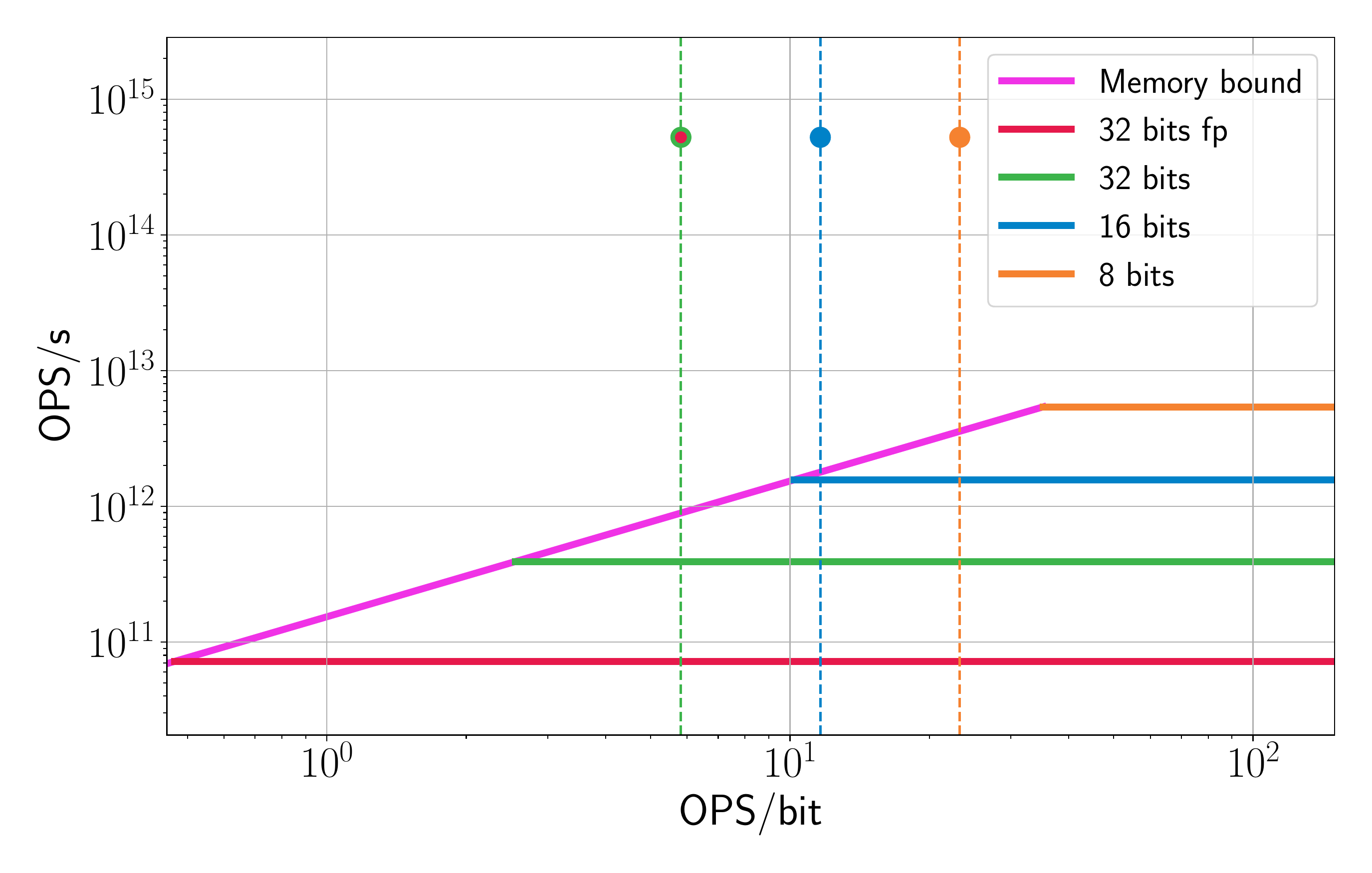}
        \caption{OPS roofline: $3\times 3$ kernel, $256$ input and output $14 \times 14$ features, $1$mm$^2$ accelerator with $800$MHz frequency, with DDR of $2.4GHz$ with $64$ bit data bus.}
        \label{fig:ops_roofline1}
\end{figure}

To calculate the amount of OPS/s required by the layer, under the assumption that a full single pixel is produced every clock, we need to calculate the amount of MAC operations required to calculate one output pixel ($n\times m \times (k^2+1)$) and  multiply it by the accelerator frequency. To calculate the OPS/bit for each design, we divide the amount of MAC operations in the layer by the total number of bits transferred over the memory bus, which includes the weights, the input and the output activations. The layer requires $\mathbf{524.288\:\nicefrac{\mathrm{TOPS}}{\mathrm{s}}}$ to be calculated without stalling for memory access and computation. The available performance of the accelerators is summarized in \cref{table:roofline_ex1} and visualised using the proposed OPS-based roofline analysis in \cref{fig:ops_roofline1}. 

\begin{table}
	\centering
	\caption{The amount of computation (OPS/s) provided by the accelerators and memory throughput (OPS/bit) required by the 11$^{th}$ layer of ResNet-18.  }
	\label{table:roofline_ex1}
	\begin{center}
		\begin{small}
			\begin{tabular}{lcccc}
				\toprule
				& \bf{32-bit} & \bf{32-bit} & \bf{16-bit} & \bf{8-bit} \\
				& \bf{float} & \bf{fixed} & \bf{quant.} & \bf{quant.}  \\
				\midrule
				GOPS/s &  $72.00$ &$392.0$& $1568$ &$5408$ \\
				OPS/bit & $5.82$&  $5.82$& $11.63$ & $23.26$ \\
				\bottomrule
			\end{tabular}
		\end{small}
	\end{center}
\end{table}

In this example, the application's requirements are out of the scope of the product definition. On one hand, all accelerators are computationally bound (all horizontal lines are below the application's requirements), indicating that we do not have enough PEs to calculate the layer in one run.  On the other hand, even if we decide to increase the computational density by using stronger quantization or by increasing the silicon area (and the cost of the accelerator), we would still hit the memory bound (represented by the diagonal line). In this case, the solution should be found at the algorithmic level or by changing the product's targets; e.g., we can calculate the layer in parts, increase the silicon area of  while decreasing the frequency in order not to hit memory wall, or decide to use another algorithm.

\begin{figure}
 \centering
        \includegraphics[width=1.0\linewidth]{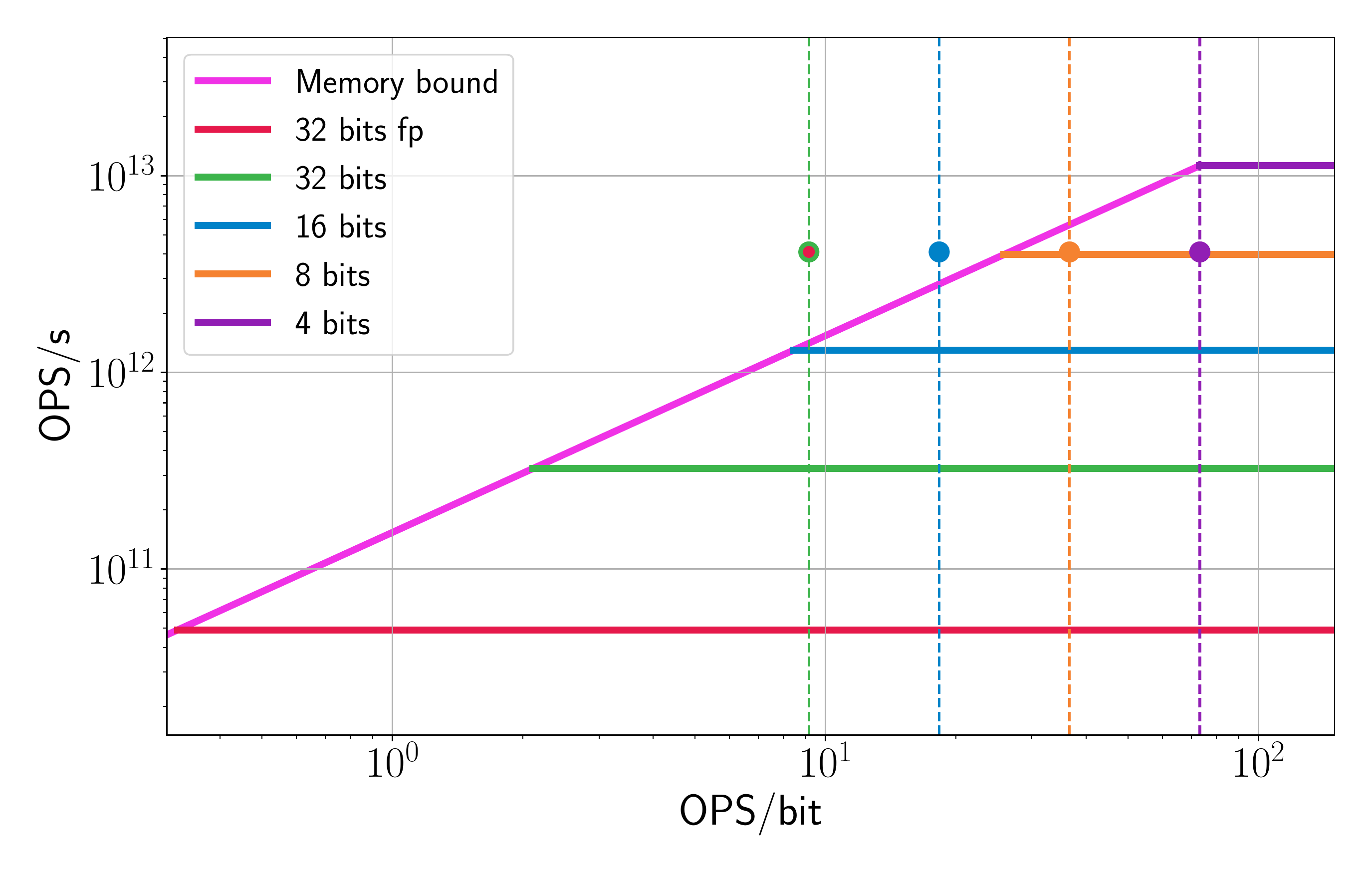}
        \caption{OPS roofline: $3\times 3$ kernel, $64$ input and output $56 \times 56$ features, $6$mm$^2$ accelerator with $100$MHz frequency, with DDR of $2.4$GHz with $64$ bit data bus.}
        \label{fig:ops_roofline2}
\end{figure}

Our second example explores the feasibility of implementing the second layer of ResNet-18 that has a $3\times3$ kernel and 64 input and output features of dimension $56\times 56$. For this example, we increase the silicon area to $\mathrm{6mm^2}$ and lower the frequency to $100$MHz, as proposed earlier, and add a 4-bit quantized accelerator for comparison purposes. The layer requires $\mathbf{4.1\:\nicefrac{\mathrm{GOPS}}{\mathrm{s}}}$. The accelerators results are summarized in \cref{table:roofline_ex2} and visualised with the OPS-based roofline analysis in \cref{fig:ops_roofline2}. 

\begin{table}
	\centering
	\caption{The amount of computation (OPS/s) provided by the accelerators and memory throughput (OPS/bit) required by second layer of ResNet-18.}
	\label{table:roofline_ex2}
	\begin{center}
		\begin{small}
			\begin{tabular}{lccccc}
				\toprule
				& \bf{32-bit} & \bf{32-bit} & \bf{16-bit} & \bf{8-bit} & \bf{4-bit}\\
				& \bf{float} & \bf{fixed} & \bf{quant.} & \bf{quant.} & \bf{quant.} \\
				\midrule
				GOPS/s &  $49.00$& $324.0$& $1296$ & $3969$& $11236$ \\
				OPS/bit & $9.16$&   $9.16$&  $18.32$& $36.64$& $73.27$\\
				\bottomrule
			\end{tabular}
		\end{small}
	\end{center}
\end{table}

From \cref{fig:ops_roofline2} we can see that our 32-bit and 16-bit accelerators are still computationally bound, while the 8-bit and 4-bit quantized accelerators meet the demands of the layer. In particular, the 8-bit accelerator is located at the border of computational ability, meaning this solution has nearly optimal resource allocation, since the hardware is fully utilized. Still, the final choice of the configuration depends on other parameters such as the accuracy of the CNN.

Both examples demonstrate that decisions made at early stages have a critical impact on the quality of the final product. For example, applying an aggressive quantization to the network or increasing the silicon size may not improve the overall performance of the chip if its performance is memory-bound. 
From the architect's point of view, it is important to balance between the computation and data transfer. Nonetheless, this balance can be achieved in different ways: at the micro-architecture level, at the algorithmic level or by changing the data representation. The architect may also consider (1) changing the hardware to provide faster communication (requires more power and is more expensive), (2) appling communication bandwidth compression algorithms \cite{baskin2019cat,chmiel2019feature}, (3) using fewer number of bits to represent weights and activations (using 3- or 4-bit representation may solve the communication problem, at the cost of reducing the expected accuracy), or (4) changing the algorithm to transfer data slower (even though that solves the bandwidth issue, the possible drawback is a reduced throughput of the whole system). The proposed OPS-based roofline model helps the architect to choose alternative. After making major architectural decisions we can use BOPS to get an estimation of the impact of different design choices on the final product, such as the expected area, power, optimal operational point, etc. 

The next section will examine these design processes from the system design point of view.

\section{HCM Metric evaluation}
\label{sec:expirements}
After introducing the use of BOPS as a metric for the hardware complexity of CNN-based algorithms and the use of the OPS-based roofline model to help the architect  understand how the decisions at the algorithmic level may impact the characterizations of the final product, this section aims to provide a holistic view of the design process of systems with CNN accelerators. We conducted an extensive evaluation of the design and the implementation of a commonly used CNN architecture for ImageNet \cite{ILSVRC15} classification, ResNet-18 \cite{7780459}. We also compared our metric to prior art \cite{mishra2018wrpn} in terms of correspondence between complexity score to hardware utilization for CNN parameters with various bitwidths.

\subsection{Experimental methodology}
\begin{figure}[h]
 \centering
        \includegraphics[width=0.6\linewidth]{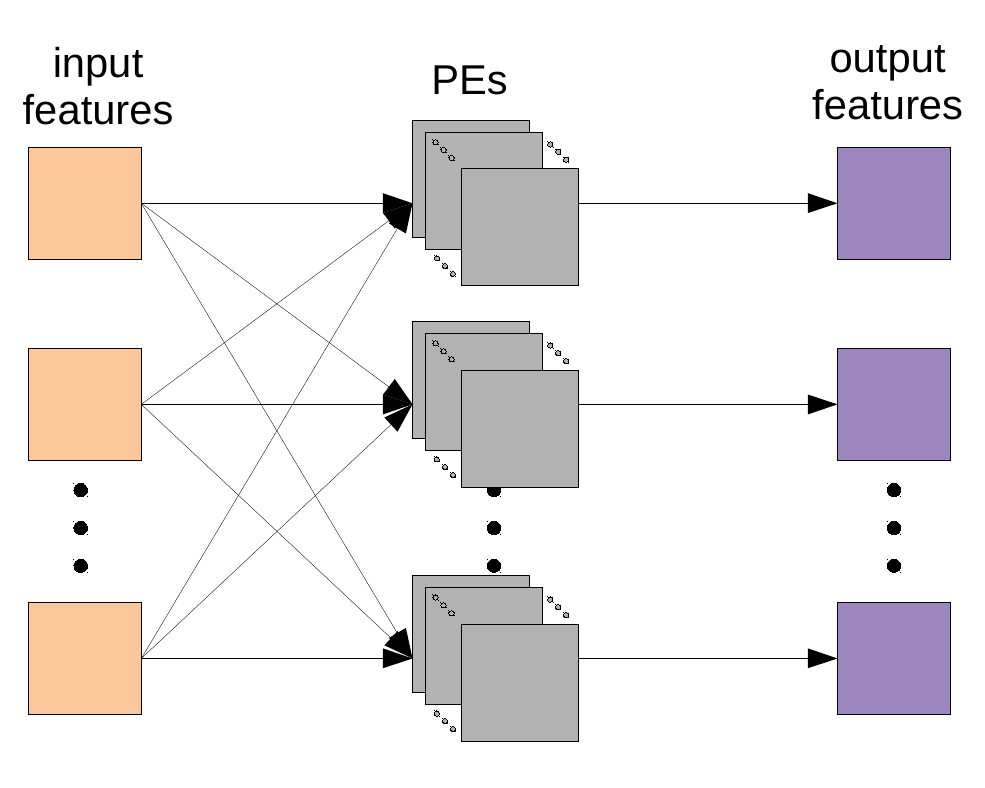}
        \caption{All-to-all topology with $n\cross m$ processing elements.}
        \label{fig:nm_topology}
\end{figure}

We start the evaluation of the HCM metric with a comprehensive review of the use of BOPS as part of the design and implementation process of a CNN accelerator. This section shows the trade-offs involved in the process and verifies the accuracy of the proposed model.
It focuses on the implementation of a single PE since PEs are directly affected by the quantization process. The area of an individual PE depends on the chosen bitwidth, while the change in the amount of input and output features changes both the required number of PEs and the size of the accumulator. The leading example  we use implemented  an all-to-all 
CNN accelerator that can calculate $n$ input features and $m$ output features in parallel, as depicted in \cref{fig:nm_topology}. For simplicity, we choose an equal number of input and output features. In this architecture, all the input features are routed to 
each of the $m$ blocks of PEs, each calculating a single output feature.
The implementation was done for an ASIC using the TSMC 28nm technology library, $800$MHz system clock and in the nominal corner of $\mathrm{V_{DD}}=$ 0.81V. For the power analysis, input activity factor, and sequential activity factor, we used the value of $0.2$. The tool versions are listed in \cref{table:design_tools}.
\begin{table}[h]
	\centering
	\caption{CAD Design Tools}
	\label{table:design_tools}
	\begin{center}
		\begin{small}
			\begin{tabular}{ll}
				\toprule
				Language & Verilog HDL \\
				Logic Simulation & ModelSim 19.1 \\
				Synthesis & Synopsys Design Compiler 2017.09-SP3 \\
				Place and route & Cadence Innovus 2019.11 \\
				\bottomrule
			\end{tabular}
		\end{small}
	\end{center}
\end{table}

For brevity, we present only the results of experiments at $800$ MHz clock frequency. We performed additional experiments at $600$ MHz and $400$ MHz (obviously, neither BOPS nor the
area of an accelerator depends on the chip frequency), but do not show these results. As shown in \cref{sec:roofline}, lowering the frequency of the design can help to avoid the memory bound, but incurs the penalty of slower solution time. 

Our results show a high correlation between the area of the design and BOPS. The choice of an all-to-all topology shown in \cref{fig:nm_topology} was made because of an intuitive understanding of how the accelerator calculates the outputs of the network. This choice, however, has a greater impact on the layout's routing difficulty, with various alternatives such as broadcast or systolic topologies \cite{chen2019eyerissv2}. For example, a systolic topology, a popular choice for high-end NN accelerators \cite{jouppi2017datacenter}, eases the routing complexity by using a mesh architecture. Although it reduces the routing effort and improves the flexibility of the input/output feature count, it requires a more complex control for the data movement to the PEs. 

\begin{figure}[h]
 \centering
        \includegraphics[width=1.0\linewidth]{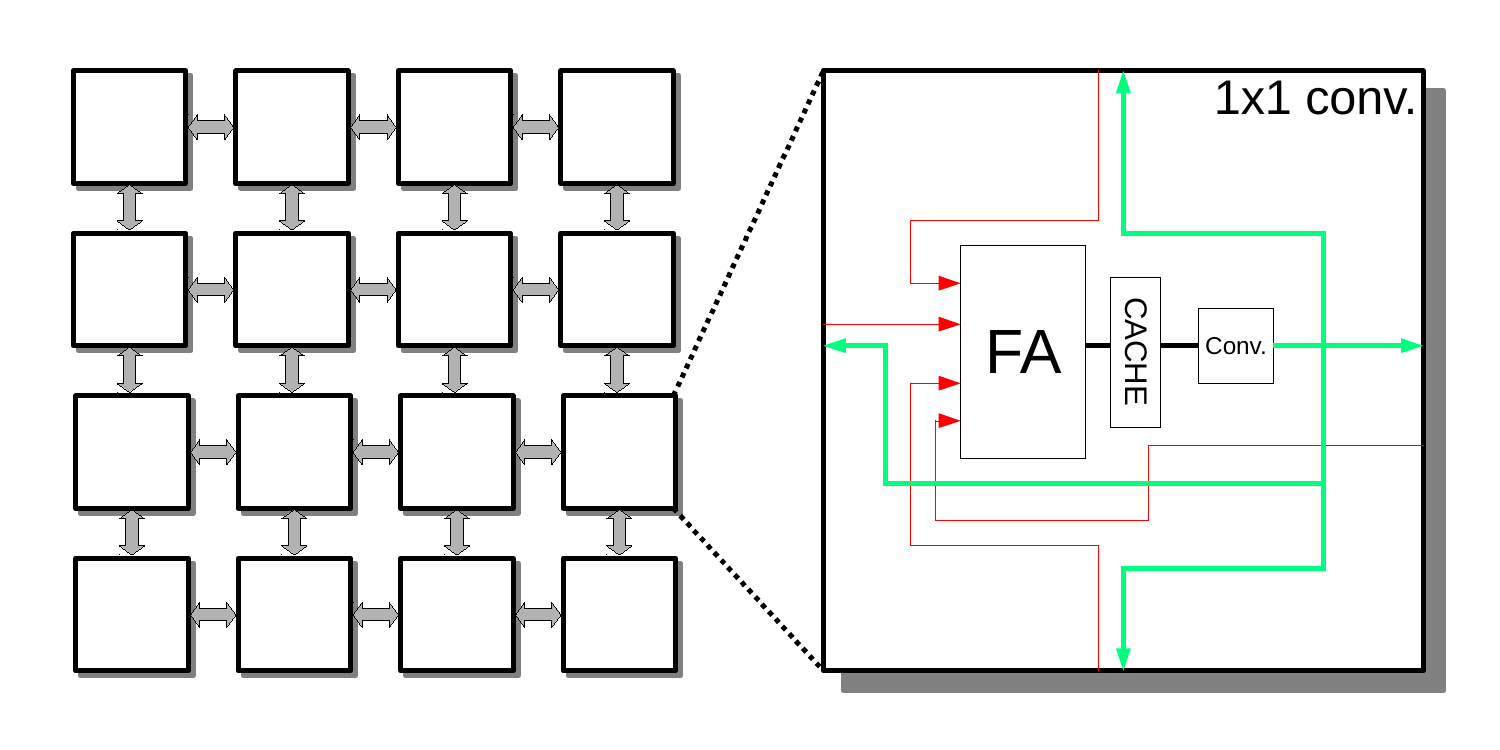}
        \caption{Systolic array of processing elements}
        \label{fig:grid_topology}
\end{figure}

To verify the applicability of BOPS to different topologies, we also implemented a systolic array shown in \cref{fig:grid_topology}, where each $1\times1$ PE is connected to 4 neighbors with the ability to bypass any input to any output without calculations. The input feature accumulator is located at the input of the PE. This topology generates natural $4\times1$ PEs, but with  proper control, it is possible to create flexible accelerators.
In the systolic design, we generated three square arrays, $4\times4$, $8\times8$, and $16\times16$, with $b_w=b_a \in \qty{4, 6}$. The systolic array area was found to be in linear relation with BOPS, with the goodness of fit $R^2=0.9752$. 

\begin{figure}
 \centering
        \includegraphics[width=1.0\linewidth]{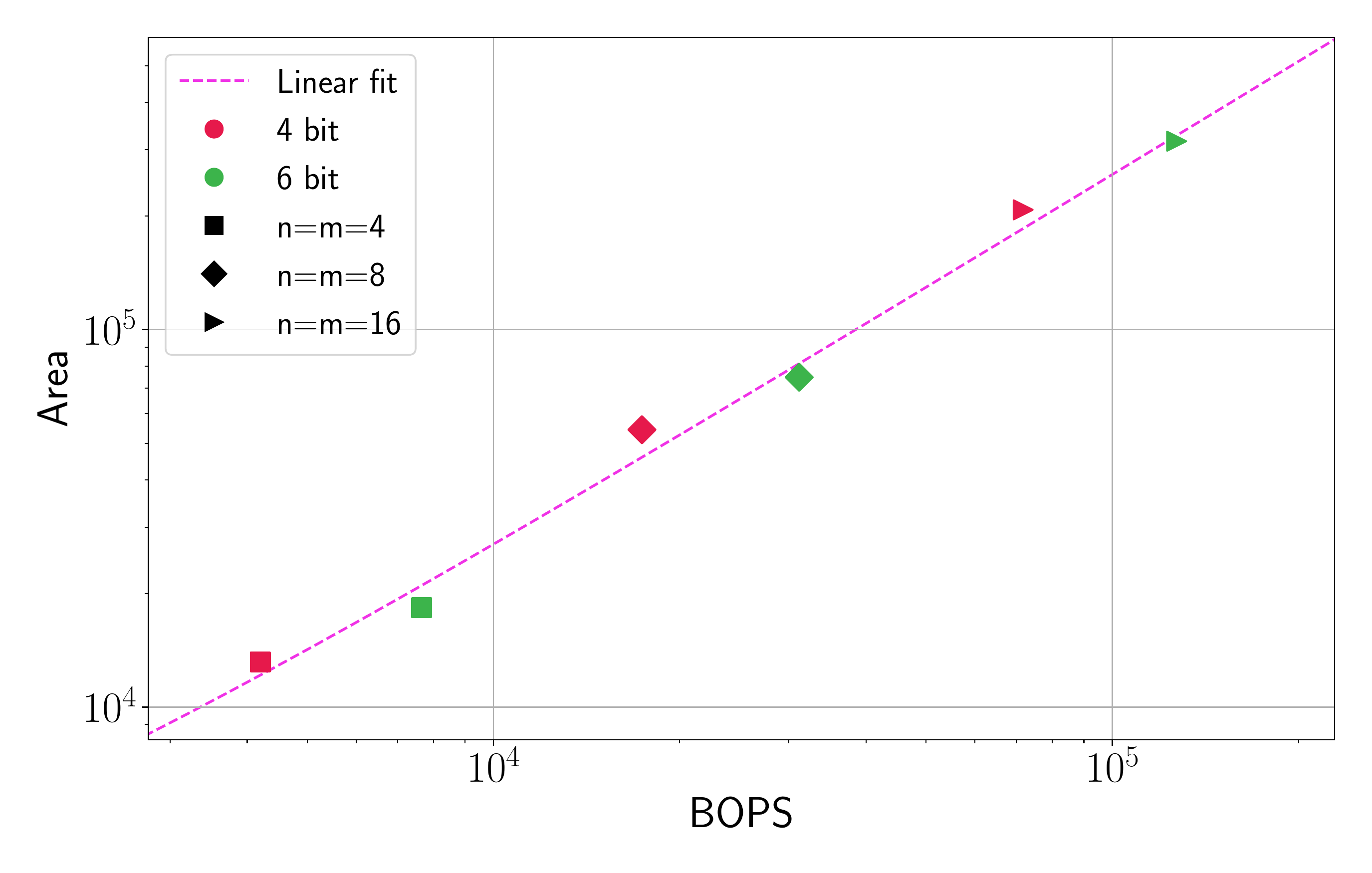}
        \caption{Area ($A$) vs. BOPS ($B$) for a systolic array of $3 \times 3$ PEs with variable input ($n$) and output ($m$) feature dimensions, and variable bitwidth. Weights and activations use the same bitwidth and the accumulator width is set to $\log_2(9m) \cdot b_w \cdot b_a$.}
        \label{fig:area_bops_systolic}
\end{figure}

\subsection{System-level design using HCM}
\begin{figure*}
 \centering
    \includegraphics[width=\textwidth]{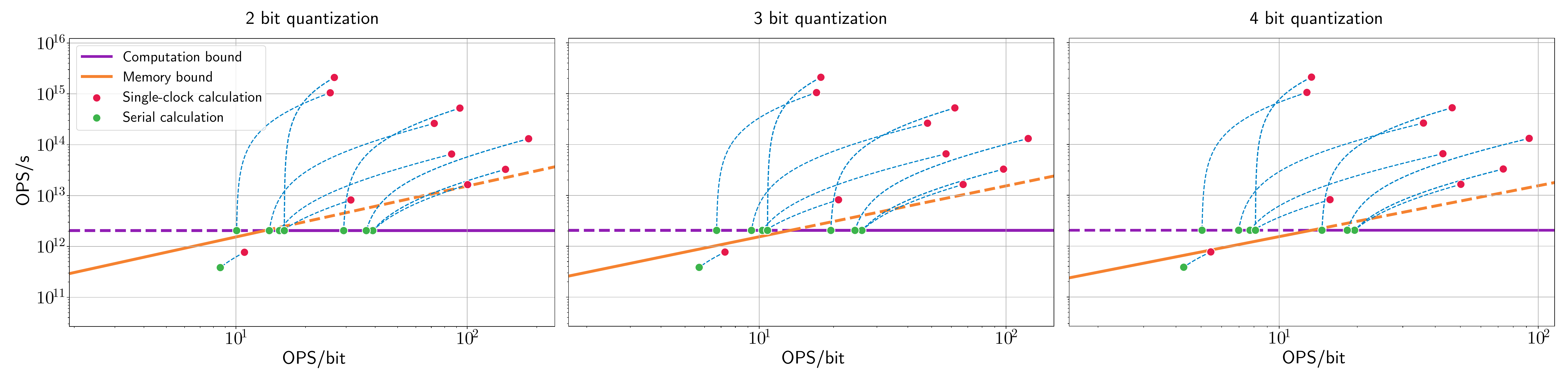}
        \caption{ResNet-18 roofline analysis for all layers. Red dots are the  performance required by the layer, and green dots are the equivalent performance using partial-sum computation. The curves connecting between the dots are linear segments distorted by the log-log axes, and are displayed only for convenience. 
        }
        \label{fig:roffline_resnet}
\end{figure*}
In this section, we analyze the acceleration of ResNet-18 using the proposed metrics and show the workflow for early estimation of the hardware cost when designing an accelerator. We start the discussion by targeting an ASIC that runs at $800$MHz, with $16\times16$ PEs and the same $2.4$GHz DDR-4 memory with a 64-bit data bus as used in \cref{sec:roofline}. 
The impact of changing these constraints is discussed at the end of the section. 
For the first layer, we replace the $7\times 7$ convolution with three $3\times 3$ convolutions, as proposed by \citet{he20199bagoftricks}. This allows us to simplify the analysis by employing universal $3\times 3$ PEs for all layers.  

We start the design process by comparing different alternatives using the new proposed OPS-based-roofline analysis since it helps to explore the design trade-offs between the multiple solutions. We calculate the amount of OPS/s provided by $16\times16$ PEs at $800$MHz and the requirements of each layer. To acquire the roofline, we need to calculate the OPS/bit, which depend on the quantization level. For ResNet-18, the current art \cite{gong2019differentiable}  achieves 69.56\% top-1 accuracy on ImageNet for 4-bit weights and activations, which is only 0.34\% less than 32-bit floating-point baseline (69.9\%). Thus we 
decided to focus on 2-, 3- and 4-bit quantization both for weights and activations, which can achieve 65.17\%, 68.66\%, and 69.56\% top-1 accuracy, correspondingly. 

For a given bitwidth, OPS/bit is calculated by dividing the total number of operations by the total number of bits transferred over the memory bus, consisting of reading weights and input activations and writing output activations. \cref{fig:roffline_resnet} presents OPS-based roofline for each quantization bitwidth. Please note that for each layer we provided two points:  the red dots are the performance required by the layer, and the green dots are the
equivalent performance using partial-sum computation.

\cref{fig:roffline_resnet} clearly indicates that this accelerator is severely limited by both computational resources and lack of enough bandwidth; the system is computationally bounded, which could be inferred from the fact that it does not have enough PEs to calculate all the features simultaneously. Nevertheless, the system is also memory-bound for any quantization level, meaning that adding more PE resources would not solve the problem. It is crucial to make this observation at the early stages of the design since it means that micro-architecture changes would not be sufficient to solve the problem.

One possible solution, as presented in \cref{sec:roofline}, is to divide the channels of the input and output feature maps into smaller groups, and use more than one clock cycle to calculate each pixel. In this way, the effective amount of the OPS/s required for the layer is reduced. In the case that the number of feature maps is divisible by the number of available PEs, the layer will fully utilize the computational resources, which is the case for every layer except the first one. Reducing the number of PEs, however, also reduces the data efficiency, and thus the OPS/bit also decreases, shifting the points to the left on the roofline plot. 
Thus, some layers still require more bandwidth from the memory than what the latter can supply. In particular, in the case of 4-bit quantization, most of the layers are memory-bound. The only option that properly utilizes the hardware is 2-bit quantization, for which all the layers except one are within the memory bound of the accelerator. 
Another option for solving the problem is to either change the neural network topology being used, or add a data compression scheme on the way to and from the memory \cite{baskin2019cat,chmiel2019feature}. Adding compression will reduce the effective memory bandwidth requirement and allow adding more PEs in order to meet the performance requirements -- at the expense of cost and power.

At this point, BOPS can be used to estimate the power and the area of each  alternative for implementing the the accelerator  using the PE micro-design. 
In addition, we can explore other trade-offs, such as the influence of modifying some parameters that were fixed at the beginning: lowering the ASIC frequency will decrease the computational bound, which reduces the cost and only hurts the performance if the network is not memory-bounded. An equivalent alternative is to decrease the number of PEs. Both procedures will reduce the power consumption of the accelerator as well the computational performance.
The system architect may also consider changing the parameters of the algorithm being used, e.g., change the feature sizes, use different quantization for the weights and for the activations, include pruning, and more. 

It is also possible to reverse design  order: start with a BOPS estimate of the number of PEs that can fit into a given area, and then calculate the ASIC frequency and memory bandwidth that would allow full utilization the accelerator. This can be especially useful if the designer has a specific area or power goal.

To summarize this section, from the architecture point of view it is extremely important to be able to predict, at the early stages of the design, if the proposed (micro)architecture is going to meet the project targets. At the project exploration stage, the system architect has plenty of alternatives to choose from  to make the right trade-offs (or even negotiate to change the product definition and  requirements. Introducing such alternatives later may be painful or even near to impossible.

\subsection{Comparison with prior metrics}
In this section, we compare the BOPS \cite{baskin2018uniq} metric to another complexity metric, introduced by \citet{mishra2018wrpn}. A good complexity metric should have a number of properties. First, it should reflect the real cost of the design. Second, it should be possible to calculate it from micro-designs or prior design results, without needing to generate complete designs. Last, it should generalize well, providing meaningful predictions for a wide spectrum of possible design parameter values. We compare our choice of computational complexity assessment, BOPS, with the ``compute cost'' proposed by \citet{mishra2018wrpn}. To analyze the metrics, we use our real accelerator area results from \cref{sec:expirements} and error bands of linear extrapolation of the measured values. To remind the reader, BOPS and  ``compute cost'' are defined as follows:
\begin{align}
    \mathrm{BOPS} &= m n k^2 \qty(b_a b_w + b_a+b_w+\log_2(n k^2)) \label{eq:BOPS} \\
    \mathrm{compute\_cost}&=m n k^2(b_a+b_w)  \label{eq:compute_cost}
\end{align}

\begin{figure}
 \centering
        \includegraphics[width=1.0\linewidth]{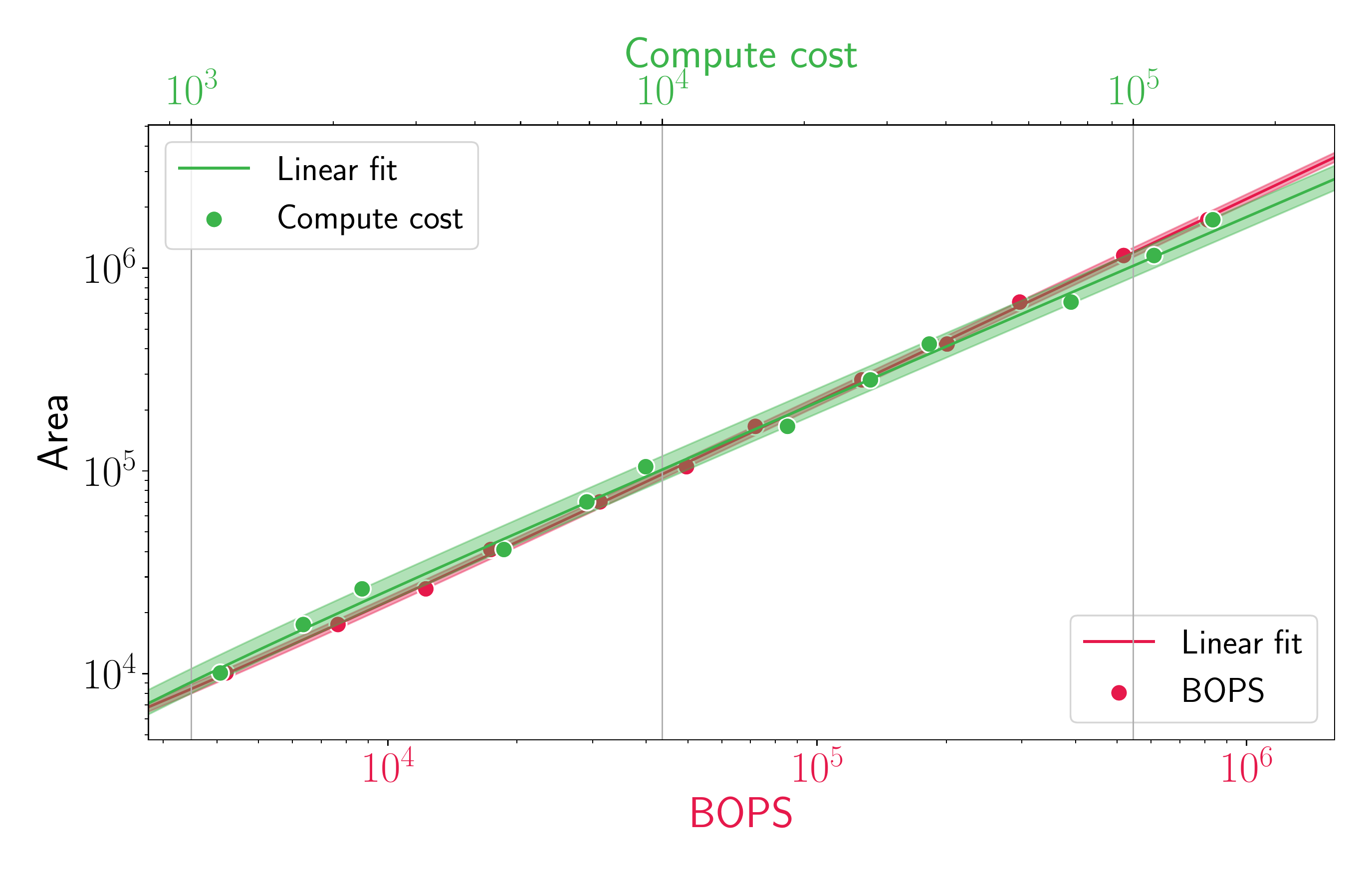}
        \caption{Comparison of BOPS and ``compute cost'' \cite{mishra2018wrpn} predictive power. BOPS -- 5\% error, ``compute cost'' -- 15\%.  }
        \label{fig:bops_vs_cc}
\end{figure}
The error of predicting a new point with ``compute cost'' is 15\% within 2 orders of magnitude, whereas using BOPS, is only 5\%. As shown in \cref{fig:bops_vs_cc}, ``compute cost'' introduces a systematic error: each of the distinguishable groups of three points corresponding to a single value of the number of input and output features creates a separate prediction line. This may lead to higher errors in case of extrapolation from a single value of the number of input and output features or a wide range of the considered bitwidth.
\section{Discussion and Conclusions}
\label{sec:conclussions}

CNN accelerators are commonly used in different systems, starting from IoT and other resource-constrained devices, and ending in datacenters and high-performance computers. Designing accelerators that meet tight constraints is still a challenging task, since the current EDA and design tools do not provide enough information to the architect. To make the right choice, the architects need to understand at the early stages of the design the impact of high-level decisions they make on the final product, and to be able to make a fair comparison between different design alternatives. 

In this paper, we showed that one of the fundamental shortcomings of the current design methodologies and tools is the use of GFLOPS as a metric for estimating the complexity of existing hardware solutions. 
The first contribution of this paper is the definition of the HCM as a metric for hardware complexity. We demonstrated its application to the prediction of  such product characteristics as power, performance, etc.

The second contribution of the paper is the introduction of the OPS-based roofline model as a supporting tool for the architect at the very early stages of the development. We showed that this model allows the comparison of different alternatives of the design and the determination of the optimality and feasibility of the solution.

Lastly, we provided several examples of realistic designs, using an actual implementation with standard design tools and a mainstream process technology. By applying the proposed metric, we could build a better system and indicate to the system architect that certain CNN architectures may better fit the constraints of a specific platform. In particular, our metric confirmed that CNN accelerators are more likely to be memory, rather that computationally bound \cite{wang2019lutnet, jouppi2017datacenter}.

Although this paper is mainly focused on ASIC-based architectures, the same methodology can be applied to many other systems, including FPGA-based implementations and other system-specific domains that allow trading off accuracy and data representation with different physical parameters such as power, performance, and area.

\subsubsection*{Acknowledgments}
The research was funded by the Hyundai Motor Company through the HYUNDAI-TECHNION-KAIST Consortium, National Cyber Security Authority, and the Hiroshi Fujiwara Technion Cyber Security Research Center.


\bibliographystyle{IEEEtranSN}
{\small
\bibliography{refs}}

\end{document}